\documentclass[11pt,a4paper,table]{article}
\usepackage[hyperref]{emnlp2019}
\usepackage{times}
\usepackage{latexsym}
\usepackage{multicol}
\usepackage{graphicx}
\usepackage{tabularx}
\usepackage{booktabs}
\usepackage{adjustbox}
\usepackage{url}

\newcolumntype{Y}{>{\centering\arraybackslash}X}

\usepackage{microtype}

\aclfinalcopy % Uncomment this line for the final submission
\title{Efficient Intent Detection with Dual Sentence Encoders \\
        \ttfamily \fontsize{11.5pt}{11.5pt}\selectfont \url{github.com/PolyAI-LDN/polyai-models}
    }
\author{
 I{\~{n}}igo Casanueva\thanks{{ } Equal contribution. TT is now at the Oxford University.},
 Tadas Tem\v{c}inas,$^{*}$
 Daniela Gerz,
 Matthew Henderson, \textmd{and}
 Ivan Vuli\'{c} \\
 PolyAI Limited \\
 London, United Kingdom \\
 \texttt{\{inigo,dan,matt,ivan\}@poly-ai.com}
}

\date{}

\begin{document}
\maketitle
\begin{abstract}
%(main point: everyone is doing BERT, however that is complicated, takes up so much computing on TPU, however you can actually get the same SOA performance with dot-product models)

Building conversational systems in new domains and with added functionality requires resource-efficient models that work under low-data regimes (i.e., in few-shot setups). Motivated by these requirements, we introduce intent detection methods backed by pretrained dual sentence encoders such as USE and ConveRT. We demonstrate the usefulness and wide applicability of the proposed intent detectors, showing that: \textbf{1)} they outperform intent detectors based on fine-tuning the full BERT-Large model or using BERT as a fixed black-box encoder on three diverse intent detection data sets; \textbf{2)} the gains are especially pronounced in few-shot setups (i.e., with only 10 or 30 annotated examples per intent); \textbf{3)} our intent detectors can be trained in a matter of minutes on a single CPU; and \textbf{4)} they are stable across different hyperparameter settings. In hope of facilitating and democratizing research focused on intention detection, we release our code, as well as a new challenging single-domain intent detection dataset comprising 13,083 annotated examples over 77 intents.

\end{abstract}

\section{Introduction}
\label{s:intro}
%\input{01-intro.tex}

% General introduction to the task and why it's important 
%(intent detection is vital for chatbot development, however there is no or little task-specific data, hence low data setting is important to solve; check blogpost)
%(OOD is important but ususally ignored)

% IV: P1 - Conversational systems are important
Task-oriented conversational systems allow users to interact with computer applications through conversation in order to solve a particular task with well-defined semantics, such as booking restaurants, hotels and flights \cite{Hemphill:1990,Williams:2012b,ElAsri:2017sigdial}, providing tourist information \cite{Budzianowski:2018emnlp}, or automating customer support \cite{Xu:2017chi}.

%% Add this if the first paragraph is too short 
%%They also provide the foundation of intelligent virtual assistants such as Amazon Alexa, Apple Siri, or Google Assistant \cite{Yu:2019emnlp}. Moreover, they are used to reduce operating costs for online customer support services \cite{Xu:2017chi}, and have found application in language learning \cite{Raux:2003,Chen:2017survey}, entertainment \cite{Fraser:2018iva}, or healthcare \cite{Laranjo:2018}.

% IV: P2 - Define intent detection, and problems with scaling and deploying quickly
\textit{Intent detection} is a vital component of any task-oriented conversational system \cite{Hemphill:1990,Coucke:18}. In order to {understand the user's current goal}, the system must leverage its intent detector to classify the user's utterance (provided in varied natural language) into one of several predefined classes, that is, \textit{intents}.\footnote{For instance, in the e-banking domain intents can be \textit{lost card} or \textit{failed top-up} (see Table~\ref{tab:banking-examples}). The importance of intent detection is also illustrated by the fact that getting the intent wrong is the first point of failure of any conversational agent.} Scaling intent detectors (as well as conversational systems in general) to support new target domains and tasks is a very challenging and resource-intensive process \cite{Wen:17,rastogi2019towards}. The need for expert domain knowledge and domain-specific labeled data still impedes quick and wide deployment of intent detectors. In other words, one crucial challenge is enabling effective intent detection in \textit{low-data scenarios} typically met in commercial systems, with only several examples available per intent (i.e., the so-called \textit{few-shot learning setups}).

%% P3 - Transfer learning, its current usage, and why we need to improve this
Transfer learning on top of pretrained sentence encoders \cite[\textit{inter alia}]{Devlin:2018arxiv,Liu:2019roberta} has now established as the mainstay paradigm aiming to mitigate the bottleneck with scarce in-domain data. However, directly applying the omnipresent sentence encoders such as BERT to intent detection may be sub-optimal. \textbf{1)} As shown by \newcite{henderson2019convert}, pretraining on a general language-modeling (LM) objective for conversational tasks is less effective than \textit{conversational pretraining} based on the response selection task \cite{Henderson:2019acl}. \textbf{2)} Fine-tuning BERT and its variants is very resource-intensive as it assumes the adaptation of the full large model. Moreover, in few-shot setups fine-tuning may result in overfitting. From a commercial perspective, these properties lead to extremely slow, cumbersome, and expensive development cycles.

Therefore, in this work we propose to use efficient \textit{dual sentence encoders} such as Universal Sentence Encoder (USE) \cite{Cer:2018arxiv} and ConveRT \cite{henderson2019convert} to support intent detection. These models are in fact neural architectures tailored for modeling sentence pairs \cite{Henderson:2019acl,Humeau:2019arxiv}, and are trained on a conversational response selection task. As such, they inherently encapsulate conversational knowledge needed for (few-shot) intent detection. We discuss their advantage over LM-based encoders, and empirically validate the usefulness of conversational pretraining for intent detection. We show that intent detectors based on fixed USE and ConveRT encodings outperform BERT-backed intent detectors across the board on three diverse intent detection datasets, with prominent gains especially in few-shot scenarios. Another advantage of dual models is their compactness:\footnote{For instance, ConveRT is only 59MB in size, pretrained in less than a day on 12 GPUs \cite{henderson2019convert}.} we demonstrate that our state-of-the-art USE+ConveRT intent detectors can be trained even on a regular laptop's CPU in several minutes. 

We also show that intent detectors based on dual sentence encoders are largely invariant to hyperparameter changes. This finding is extremely important for real-life low-data regimes: due to the invariance, the expensive hyperparameter tuning step can be bypassed, and a limited number of annotated examples can be used directly as additional training data, instead of held-out validation data.

%% P5 - Another contribution, dataset
Another contribution of this work is a new and challenging intent detection dataset in the banking domain, dubbed \textsc{banking77}. It follows the very recent endeavor of procuring high-quality intent detection data \cite{Liu:2019iwsds,larson-etal-2019-evaluation}, but is very different in nature than the other datasets. Unlike prior work which scatters a set of coarse-grained intents across a multitude of domains (i.e., 10+ domains, see Table~\ref{tab:data} later), we present a challenging single-domain dataset comprising 13,083 examples over 77 fine-grained intents. We release the code and the data online at: \\
{\small \url{github.com/PolyAI-LDN/polyai-models}}.

\section{Methodology: Intent Detection with Dual Sentence Encoders}
\label{s:methodology}

\textbf{Pretrained Sentence Encoders.} Large-scale pretrained models have benefited a wide spectrum of NLP applications immensely \cite{Devlin:2018arxiv,Liu:2019roberta,radford2019language}. Their core strength lies in the fact that, through consuming large general-purpose corpora during pretraining, they require smaller amounts of domain-specific training data to adapt to a particular task and/or domain \cite{Ruder:2019transfer}. The adaptation is typically achieved by adding a task-specific output layer to a large pretrained sentence encoder, and then fine-tuning the entire model \cite{Devlin:2018arxiv}. However, the fine-tuning process is computationally intensive \cite{Zafrir:2019arxiv,henderson2019convert}, and still requires sufficient task-specific data \cite{Arase:2019emnlp,Sanh:2019arxiv}. As such, the standard approach is both unsustainable in terms of resource consumption \cite{Strubell:2019acl}, as well as sub-optimal for few-shot scenarios.

%as it requires substantially larger amounts of data.

\vspace{1.6mm}
\noindent \textbf{Dual Sentence Encoders and Conversational Pretraining.} A recent branch of sentence encoders moves beyond the standard LM-based pretraining objective, and proposes an alternative objective: \textit{conversational response selection}, typically on Reddit data \cite{AlRfou:2016arxiv,Henderson:2019arxiv}. As empirically validated by \newcite{Henderson:2019acl,Mehri:2019acl}, conversational (instead of LM-based) pretraining aligns better with conversational tasks such as dialog act prediction or next utterance generation. 

Pretraining on response selection also allows for the use of efficient \textit{dual} models: the neural response selection architectures are instantiated as dual-encoder networks that learn the interaction between inputs/contexts and their relevant (follow-up) responses. Through such response selection pretraining regimes they organically encode useful conversational cues in their representations. %For more technical details related to response selection pretraining we refer the interested reader to prior work \cite{Cer:2018arxiv,henderson2019convert}.

In this work, we propose to use such efficient conversational dual models as the main source of (general-purpose) conversational knowledge to inform domain-specific intent detectors. We empirically demonstrate their benefits over other standard sentence encoders such as BERT in terms of \textbf{1)} performance, \textbf{2)} efficiency, and \textbf{3)} applicability in few-shot scenarios. We focus on two prominent dual models trained on the response selection task: Universal Sentence Encoder (USE) \cite{Cer:2018arxiv}, and Conversational Representations from Transformers (ConveRT) \cite{henderson2019convert}. For further technical details regarding the two models, we refer the interested reader to the original work.

\vspace{1.6mm}
\noindent \textbf{Intent Detection with dual Encoders.} 
We implement a simple yet effective model (see \S\ref{s:results} later) for intent detection which is based on the two dual models. Unlike with BERT, we do not fine-tune the entire model, but use fixed sentence representations encoded by USE and ConveRT. We simply stack a Multi-Layer Perceptron (MLP) with a single hidden layer with ReLU non-linear activations \cite{Maas:2014icml} on top of the fixed representations, followed by a softmax layer for multi-class classification. This simple formulation also allows us to experiment with the combination of USE and ConveRT representations: we can feed the concatenated vectors to the same classification architecture without any further adjustment.

\section{New Dataset: \textsc{banking77}}
\label{s:banking77}
%\input{02b-dataset.tex}
%\textbf{Previous Work and Motivation.} 
In spite of the crucial role of intent detection in any task-oriented conversational system, publicly available intent detection datasets are still few and far between. The previous standard datasets such as Web Apps, Ask Ubuntu, the Chatbot Corpus \cite{Braun:17} or SNIPS \cite{Coucke:18} are limited to only a small number of classes ($<10$), which oversimplifies the intent detection task and does not emulate the true environment of commercial systems. Therefore, more recent work has recognized the need for improved intent detection datasets. \textbf{1)} The dataset of \newcite{Liu:2019iwsds}, dubbed \textsc{hwu64}, contains 25,716 examples for 64 intents in 21 domains. \textbf{2)} The dataset of \newcite{larson-etal-2019-evaluation}, dubbed \textsc{clinc150}, spans 150 intents and 23,700 examples across 10 domains. 

However, the two recent datasets are \textit{multi-domain}, and the examples per each domain may not sufficiently capture the full complexity of each domain as encountered ``in the wild''. Therefore, to complement the recent effort on data collection for intent detection, we propose a new \textit{single-domain} dataset: it provides a very fine-grained set of intents in a banking domain, not present in \textsc{hwu64} and \textsc{clinc150}. The new \textsc{banking77} dataset comprises 13,083 customer service queries labeled with 77 intents. Its focus on fine-grained single-domain intent detection makes it complementary to the two other datasets: we believe that any comprehensive intent detection evaluation should involve both coarser-grained multi-domain datasets such as \textsc{hwu64} and \textsc{clinc150}, and a fine-grained single-domain dataset such as \textsc{banking77}. The data stats are summarized in Table~\ref{tab:data}.

%% for all three datasets 

The single-domain focus of \textsc{banking77} with a large number of intents makes it more challenging. Some intent categories partially overlap with others, which requires fine-grained decisions, see Table~\ref{tab:banking-examples} (e.g., \textit{reverted top-up} vs. \textit{failed top-up}). Furthermore, as other examples from Table~\ref{tab:banking-examples} suggest, it is not always possible to rely on the semantics of individual words to capture the correct intent.\footnote{The examples in \textsc{banking77} are also longer on average (12 words) than in \textsc{hwu64} (7 words) or \textsc{clinc150} (8).}

\begin{table}[!t]
\footnotesize
\centering
\begin{tabularx}{\linewidth}{l XXX}
\toprule
\textbf{Dataset} & \textbf{Intents} & \textbf{Examples} & \textbf{Domains}\\
\midrule
{\textsc{hwu64}} & 64 & 25,716 & 21 \\
{\textsc{clinc150}} & 150 & 23,700 & 10 \\
\cmidrule(lr){2-4}
{\textsc{banking77} (ours)} & 77 & 13,083 & 1  \\
\bottomrule
\end{tabularx}
\vspace{-1mm}
\caption{Intent detection datasets: key statistics.}
\label{tab:data}
\vspace{-1.5mm}
\end{table}

\begin{table*}[!t]
\footnotesize
\def\arraystretch{0.95}
\centering
\begin{tabularx}{\linewidth}{l X}
\toprule
\textbf{Intent Class} & \textbf{Example Utterance}  \\
\midrule
{Card Lost} & \textcolor{black}{\em Could you assist me in finding my lost card?}    \\
{Link to Existing Card} & \textcolor{black}{\em I found my lost card. Am I still able to use it?} \\
{Reverted Top-up} & \textcolor{black}{\em Hey, I thought my topup was all done but now the money is gone again -- what's up with that?} \\
{Failed Top-up} & \textcolor{black}{\em Tell me why my topup wouldn't go through?}   \\
\bottomrule
\end{tabularx}
\caption{Intent classes and example utterances from \textsc{banking77}.}
\label{tab:banking-examples} 
\end{table*}

\section{Experimental Setup}
\label{s:exp}

\textbf{Few-Shot Setups.} We conduct all experiments on the three intent detection datasets described in \S\ref{s:banking77}. We are interested in wide-scale few-shot intent classification in particular: we argue that this setup most closely resembles the development process of a commercial conversational system, which typically starts with only a small number of data points when expanding to a new domain or task. We simulate such low-data settings by sampling smaller subsets from the full data. We experiment with setups where only 10 or 30 examples are available for each intent, while we use the same standard test sets for each experimental run.\footnote{For reproducibility, we release all training subsets.}

\begin{table*}[!t]
\centering
{\footnotesize
\begin{tabularx}{\linewidth}{l XXX XXX XXX}
\toprule
  {} & \multicolumn{3}{c}{\bf \textsc{banking77}} & \multicolumn{3}{c}{\bf \textsc{clinc150}} & \multicolumn{3}{c}{\bf \textsc{hwu64}} \\
  \cmidrule(lr){2-4} \cmidrule(lr){5-7} \cmidrule(lr){8-10}
\textbf{Model} & \textbf{10}  & \textbf{30} & \textbf{Full} & \textbf{10}  & \textbf{30} & \textbf{Full} & \textbf{10}  & \textbf{30} & \textbf{Full} \\
 \cmidrule(lr){2-4} \cmidrule(lr){5-7} \cmidrule(lr){8-10}
\textsc{bert-fixed} & \textcolor{black}{67.55} & \textcolor{black}{80.07} & \textcolor{black}{87.19}  &\textcolor{black}{80.16} & \textcolor{black}{87.99} & \textcolor{black}{91.79} & \textcolor{black}{72.61} & \textcolor{black}{79.78} & \textcolor{black}{85.77} \\
\textsc{bert-tuned} & 83.42 & 90.03 & \textbf{93.66}  & 91.93 & 95.49 & 96.93  & 84.86 &  88.27 & 92.10 \\
\textsc{USE} & 84.23 & 89.74 & 92.81  & 90.85 & 93.98 & 95.06  & 83.75 & 89.03 & 91.25 \\
\textsc{ConveRT} & 83.32 & 89.37 & 93.01  & 92.62 & 95.78 & \textbf{97.16}  & 82.65 & 87.88 & 91.24 \\
\textsc{USE+ConveRT} & \textbf{85.19} & \textbf{90.57} & {93.36} & \textbf{93.26} & \textbf{96.13} & \textbf{97.16}  & \textbf{85.83} & \textbf{90.16} & \textbf{92.62} \\
\bottomrule
\end{tabularx}}%
\caption{Accuracy scores ($\times$100\%) on all three intent detection data sets with varying number of training examples (\textbf{10} examples per intent; \textbf{30} examples per intent; \textbf{Full} training data). The peak scores per column are in bold.}
\label{results-table}
\end{table*}

%% IV (this is for the discussion, not for the table):
%% Concatenating USE with ConveRT embeddings leads to the best performance throughout. We find this setup to be especially advantageous in low-data regimes.

\iffalse
\begin{table}[]
\begin{tabular}{|l|c|c|}
\hline
Encoder    & CPU & GPU \\ \hline
Bert Large & 2.4  & 235.9  \\ \hline
USE        & 53.5  & 785.4  \\ \hline
ConveRT    & 58.3  & 866.7  \\ \hline
\end{tabular}
\caption{Number of sentences encoded per second by the 3 sentence encoders benchmarked.}
\label{profiling-table}
\end{table}

\begin{table}[]
\begin{tabular}{|l|c|c|c|}
\hline
model    & CPU & GPU & TPU \\ \hline
Bert finetuned & N/A  & N/A & 567s  \\ \hline
USE        & 65s  & 57s & N/A \\ \hline
ConveRT    & 73s  & 53s & N/A \\ \hline
\end{tabular}
\caption{Time to train and evaluate an intent detection model using the USE, ConveRT and BERT-finetuned intent detection models for the dataset banking and the data regime 10. The CPU is a 2.3 GHz Dual-Core Intel Core i5. The GPU is a GeForce RTX 2080 Ti, 11GiB. The TPU is a v2-8, 8 cores, 64 GiB.}
\label{profiling-table2}
\end{table}
\fi

\vspace{1.6mm}
\noindent \textbf{MLP Design.} 
Unless stated otherwise (e.g., in experiments where we explicitly vary hyperparameters), for the MLP classifier, we use a single 512-dimensional hidden layer. We train with stochastic gradient descent (SGD), with the learning rate of $0.7$ and linear decay. We rely on very aggressive dropout ($0.75$) and train for $500$ iterations to reach convergence. We show how this training regime can improve the model's generalization capability, and we also probe its (in)susceptibility to diverse hyperparameter setups later in \S\ref{s:results}. Low-data settings are balanced, which is especially easy to guarantee in few-shot scenarios.

\vspace{1.6mm}
\noindent \textbf{Models in Comparison.} We compare intent detectors supported by the following pretrained sentence encoders. First, in the \textsc{bert-fixed} model we use pretrained BERT in the same way as dual encoders, in the so-called \textit{feature mode}: we treat BERT as a black-box fixed encoder and use it to compute encodings/features for training the classifier.\footnote{We have also experimented with ELMo embeddings \cite{Peters:2018naacl} in the same feature mode, but they are consistently outperformed by all other models in comparison.} We use the mean-pooled ``sequence ouput'' (i.e., the pooled mean of the sub-word embeddings) as the sentence representation.\footnote{This performed slightly better than using the [CLS] token embedding as sentence representation.} In the \textsc{bert-tuned} model, we rely on the standard BERT-based fine-tuning regime for classification tasks \cite{Devlin:2018arxiv} which adapts the full model. We train a softmax layer on top of the \textsc{[cls]} token output. We use the Adam optimizer with weight decay and a learning rate of $4 \times 10^{-4}$. For low-data (10 examples per intent), mid-data (30 examples) and full-data settings we train for 50, 18, and 5 epochs, respectively, which is sufficient for the model to converge, while avoiding overfitting or catastrophic forgetting.

We use the two publicly available pretrained dual encoders: \textbf{1)} the multilingual large variant of \textsc{use} \cite{Yang:2019multiuse},\footnote{https://tfhub.dev/google/universal-sentence-encoder-multilingual-large/1} and \textbf{2)} the single-context \textsc{ConveRT} model trained on the full 2015-2019 Reddit data comprising 654M \textit{(context, response)} training pairs \cite{henderson2019convert}.\footnote{https://github.com/PolyAI-LDN/polyai-models} In all experimental runs, we rely on the pretrained cased BERT-large model: 24 Transformer layers, embedding dimensionality 1024, and a total of 340M parameters. Note that e.g. ConveRT is much lighter in its design and is also pretrained more quickly than BERT \cite{henderson2019convert}: it relies on 6 Transfomer layers with embedding dimensionality of 512. We report accuracy as the main evaluation measure for all experimental runs.

\section{Results and Discussion}
\label{s:results}
\begin{table*}[!t]
\centering
\begin{adjustbox}{max width=\linewidth}
{\fontsize{7.5pt}{7.5pt}\selectfont
\begin{tabularx}{\linewidth}{l XX XX XX}
\toprule
  {} & \multicolumn{2}{c}{\bf \textsc{banking77}} & \multicolumn{2}{c}{\bf \textsc{clinc150}} & \multicolumn{2}{c}{\bf \textsc{hwu64}} \\
  \cmidrule(lr){2-3} \cmidrule(lr){4-5} \cmidrule(lr){6-7}
\textbf{Model} & \textbf{10}  & \textbf{Full} & \textbf{10}   & \textbf{Full} & \textbf{10} & \textbf{Full} \\
 \cmidrule(lr){2-3} \cmidrule(lr){4-5} \cmidrule(lr){6-7}
\textsc{bert-fixed} & {64.9 (67.8) [57.0]} & {86.2 (88.4) [74.9]} & {78.1 (80.6) [70.2]} & {91.2 (92.6) [84.7]} & {71.5 (72.8) [68.0]} & {85.9 (86.8) [81.5]} \\
\textsc{USE} & {83.9 (84.4) [83.0]} & {92.6 (92.9) [91.4]} & {90.6 (91.0) [89.9]} & {95.0 (95.3) [93.9]} & {83.6 (83.9) [83.0]} & {91.6 (92.1) [90.7]} \\
\textsc{ConveRT} & {83.1 (83.4) [82.4]} & {92.6 (93.0) [91.6]} & {92.4 (92.8) [92.0]} & {97.1 (97.2) [96.3]} & {82.5 (83.1) [82.0]} & {91.3 (91.6) [90.8]} \\
\textsc{USE+ConveRT} & {85.2 (85.5) [84.8]} & {93.3 (93.5) [92.8]} & {93.2 (93.5) [92.8]} & {97.0 (97.2) [96.5]} & {85.9 (86.2) [85.7]} & {92.5 (92.8) [91.6]} \\
\bottomrule
\end{tabularx}}%
\end{adjustbox}
\caption{Variation in accuracy scores ($\times$100\%) with different hyperparameter regimes for all the models in comparison and on all three datasets. \textbf{10} again means 10 training examples per intent as opposed to \textbf{Full} training data. The scores are provided as \textit{avg (max) [min]}: \textit{avg} is the average over all runs with different hyperparameter settings for each encoder model and each setup, \textit{max} and \textit{min} are the respective maximum and minimum scores.}
\label{results-variation}
\end{table*}

\begin{table}[!h]
{\footnotesize
\begin{tabularx}{\linewidth}{l YY}
\toprule
Encoder    & CPU & GPU \\ \cmidrule(lr){2-3}
\textsc{bert} (Large) & 2.4  & 235.9  \\ 
\textsc{USE}        & 53.5  & 785.4  \\ 
\textsc{ConveRT}    & 58.3  & 866.7  \\
\bottomrule
\end{tabularx}}%
\caption{Average number of sentences encoded \textit{per second} with the three sentence encoders. The data is fed to each encoder in batches of 15 sentences.}
\label{profiling-table}
\end{table}

\begin{table}[!h]
{\footnotesize
\begin{tabularx}{\linewidth}{l YYY}
\toprule
Classifer    & CPU & GPU & TPU \\ \cmidrule(lr){2-4}
\textsc{bert-tuned} & n/a  & n/a & 567s  \\ 
\textsc{USE}        & 65s  & 57s & n/a \\ 
\textsc{ConveRT}    & 73s  & 53s & n/a \\ 
\bottomrule
\end{tabularx}}%
\caption{Time to train and evaluate an intent classification model based on two dual models and fine-tuning \textsc{bert} on \textsc{banking77} in a few-shot scenario with 10 examples per intent. The CPU is a 2.3 GHz Dual-Core Intel Core i5. The GPU is a GeForce RTX 2080 Ti, 11 GB. The TPU is a v2-8, 8 cores, 64 GB.}
\label{profiling-table2}
\end{table}

Table~\ref{results-table} summarizes the main results; we show the accuracy scores of all models on all three datasets, and for different training data setups. As one crucial finding, we report competitive performance of intent detectors based on the two dual models, and their relative performance seems to also depend on the dataset at hand: \textsc{USE} has a slight edge over \textsc{ConveRT} on \textsc{hwu64}, but the opposite holds on \textsc{clinc150}. The design based on fixed sentence representations, however, allows for the straightforward combination of \textsc{USE} and \textsc{ConveRT}. The results suggest that the two dual models in fact capture complementary information, as the combined \textsc{USE+ConveRT}-based intent detectors result in peak performance across the board. As discussed later, due to its pretraining objective, BERT is competitive only in its fine-tuning mode of usage, and cannot match other two sentence encoders in the feature-based (i.e., fixed) usage mode. 

%%\inigo{Remember to mention that bert finetuned doesnt learn at all in 2\% of the runs, and that we removed these seeds from the final results (i guess we could call it catastrophic forgetting, but im not sure if its this or it just never starts learning)}

\vspace{1.6mm}
\noindent \textbf{Few-Shot Scenarios.} 
The focus of this work is on low-data few-shot scenarios often met in production, where only a handful of annotated examples per intent are available. The usefulness of dual sentence encoders comes to the fore especially in this setup: 1) the results indicate gains over the fine-tuned BERT model especially for few-shot scenarios, and the gains are more pronounced in our ``fewest-shot'' setup (with only 10 annotated examples per intent). The respective improvements of \textsc{USE+ConveRT} over \textsc{bert-tuned} are +1.77, +1.33, and +0.97 for \textsc{banking77}, \textsc{clinc150}, and \textsc{hwu64} (10 examples per intent), and we also see better results with the combined model when 30 examples per intent are available on all three datasets. Overall, this proves the suitability of dual sentence encoders for the few-shot intent classification task.

%%Moreover, as a result of their efficiency, it is straightforward to train dual models for many iterations. We observe that by training with a high dropout of 0.7 and many iterations (500) on a small train set \dan{INIGO Is this what you did? How long does this training take on your laptop?}, the model generalises well and also becomes highly reliable. We find under this \textit{high-dropout long-training} regime the model becomes largely insensitive to hyperparameters \dan{This needs numbers/table to back it up.}, which also removes the need of a validation set to choose the best model. Note this is not the case with BERT. Here we find outliers and low validation accuracy approximately \dan{INIGO every 3 out of X seeds}, hence it is crucial to keep a validation set to ensure a working model. \dan{have we tried the same training regime with BERT and know it doesn't work? Or we can't do it at all because it's too expensive?} 

\vspace{1.6mm}
\noindent \textbf{Invariance to Hyperparameters.}
A prominent risk in few-shot setups concerns overfitting to small data sets \cite{Srivastava:2014jmlr,Olson:2018nips}. Another issue concerns the sheer lack of training data, which gets even more pronounced if a subset of the (already scarce) data must be reserved for validation and hyper-parameter tuning. Therefore, a desirable property of any few-shot intent detector is its invariance to hyperparameters and, consequently, its off-the-shelf usage without further tuning on the validation set. This effectively means that one could use all available annotated examples directly for training. In order to increase the reliability of the intent detectors and prevent overfitting in few-shot scenarios, we suggest to use the aggressive dropout regularization (i.e., the dropout rate is 0.75), and a very large number of iterations (500), see \S\ref{s:exp}.

We now demonstrate that the intent detectors based on dual encoders are very robust with respect to different hyper-parameter choices, starting from this basic assumption that a high number of iterations and high dropout rates $r$ are needed. For each classifier, we fix the \textit{base/pivot} configuration from \S\ref{s:exp}: the number of hidden layers is $H=1$, its dimensionality is $h=512$, the SGD optimizer is used with the learning rate of $0.7$. Starting from the pivot configuration, we create other configurations by altering one hyper-parameter at the time from the pivot. We probe the following values: $r=\{0.75, 0.5, 0.25\}$, $H=\{0, 1, 2\}$, $h=\{128, 256, 512, 1024\}$, and we also try out all the configurations with another optimizer: Adam with the linearly decaying learning rate of $4 \times 10^{-4}$.

The results with all hyperparameter configs are summarized in Table~\ref{results-variation}. They suggest that intent detectors based on dual models are indeed very robust. Importantly, we do not observe any experimental run which results in substantially lower performance with these models. In general, the peak scores with dual-based models are reported with higher $r$ rates (0.75), and with larger hidden layer sizes $h$ (1,024). On the other side of the spectrum are variants with lower $r$ rates (0.25) and smaller $h$-s (128). However, the fluctuation in scores is not large, as illustrated by the results in Table~\ref{results-variation}. This finding does not hold for \textsc{bert-fixed} where in Table~\ref{results-variation} we do observe ``outlier'' runs with substantially lower performance compared to its peak and average scores. Finally, it is also important to note \textsc{bert-tuned} does not converge to a good solution for 2\% of the runs with different seeds, and such runs are not included in the final reported numbers with that baseline in Table~\ref{results-table}.

\vspace{1.6mm}
\noindent \textbf{Resource Efficiency.} 
Besides superior performance established in Table~\ref{results-table} and increased stability (see Table~\ref{results-variation}), another advantage of the two dual models is their \textit{encoding efficiency}. In Table~\ref{profiling-table} we report the average times needed by each fixed encoder to encode sentences fed in the batches of size 15 on both CPU (2.3 GHz Dual-Core Intel Core i5) and GPU (GeForce RTX 2080 Ti, 11 GB). The encoding times reveal that \textsc{bert}, when used as a sentence encoder, is around 20 times slower on the CPU and  roughly 3 times slower on the GPU.\footnote{We provide a \textit{colab} script to reproduce these experiments.}

Furthermore, in Table~\ref{profiling-table2} we present the time required to train and evaluate an intent classification model for \textsc{banking77} in the lowest-data regime (10 instances per intent).\footnote{Note that we cannot evaluate \textsc{bert-tuned} on GPU as it runs out of memory. Similar problems were reported in prior work \cite{Devlin:2018arxiv}. \textsc{USE} and \textsc{ConveRT} cannot be evaluated on TPUs as they currently lack TPU-specific code.}  Note that the time reduction on GPU over CPU for the few-shot scenario is mostly due to the reduced encoding time on GPU (see Table~\ref{profiling-table} again). However, when operating in the \textit{Full} data regime, the benefits of GPU training vanish: using a neural net with a single hidden layer the overhead of the GPU usage is higher than the speed-up achieved due to faster encoding and network computations. Crucially, the reported training and execution times clearly indicate that effective intent detectors based on pretrained dual models can be constructed even without large resource demands and can run even on CPUs, without huge models that require GPUs or TPUs. In sum, we hope that our findings related to improved resource efficiency of dual models, as well as the shared code will facilitate further and wider research focused on intent detection.

%NOTE: The take message here is that to train small intent detection models we dont need huge models that require TPUs like BERT, even if the train time is not huge. To be honest, the smaller training time of the GPU models is because the smaller encoding time, since the training time is longer (as it is only a layer, the overhead of the GPU is worse that the speed up in the matmuls). Think about how we want to present this. But this shows how good these features (the encodings of convert and USE) are to train this models.

%NOTE2: if we want to present model training results in banking full then CPU really outperforms GPU. The training time results in this case are ~8m for CPU (2m encodings 6m model training), ~11m for GPU (6s encodings 11m model training). 

\vspace{1.6mm}
\noindent \textbf{Further Discussion.}
The results from Tables~\ref{results-table} and \ref{results-variation} show that transferring representations from conversational pretraining based on the response selection task is useful for conversational tasks such as intent detection. This corroborates the main findings from prior work \cite{Humeau:2019arxiv,henderson2019convert}. The results also suggest that using BERT as an off-the-shelf sentence encoder is sub-optimal: BERT is much more powerful when used in the fine-tuning mode instead of the less expensive ``feature-based'' mode \cite{Peters:2019repl}. This is mostly due to its pretraining LM objective: while both USE and ConveRT are forced to reason at the level of full sentences during the response selection pretraining, BERT is primarily a (local) language model. It seems that the next sentence prediction objective is not sufficient to learn a universal sentence encoder which can be applied off-the-shelf to unseen sentences in conversational tasks \cite{Mehri:2019acl}. However, BERT's competitive performance in the fine-tuning mode, at least in the \textit{Full} data scenarios, suggests that it still captures knowledge which is useful for intent detection. Given strong performance of both fine-tuned BERT and dual models in the intent detection task, in future work we plan to investigate hybrid strategies that combine dual sentence encoders and LM-based encoders. Note that it is also possible to combine \textsc{bert-fixed} with the two dual encoders, but such ensembles, besides yielding reduced performance, also substantially increase training times (Table~\ref{profiling-table}).

We also believe that further gains can be achieved by increasing the overall size and depth of dual models such as ConveRT, but this comes at the expense of its efficiency and training speed: note that the current architecture of ConveRT relies on only 6 Transformer layers and embedding dimensionality of 512 (cf., BERT-Large with 24 layers and 1024-dim embeddings).

%potential footnote: BERT you need a validation set, it is less reliable. 3 out of X seeds this happens, you need to rerun
%we are not sure why, potentially many iterations because of the very high dropout %\dan{TODO} high dropout and long training works well, hyperparameters do not really matter

\section{Conclusion}
\label{s:conclusion}

We have presented intent classification models that rely on sentence encoders which were pretrained on a conversational response selection task. We have demonstrated that using dual encoder models such as USE and ConveRT yield state-of-the-art intent classification results on three diverse intent classification data sets in English. One of these data sets is another contribution of this work: we have proposed a fine-grained single-domain data set spanning 13,083 annotated examples across 77 intents in the banking domain. 

The gains with the proposed models over fully fine-tuned BERT-based classifiers are especially pronounced in few-shot scenarios, typically encountered in commercial systems, where only a small set of annotated examples per intent can be guaranteed. Crucially, we have shown that the proposed intent classifiers are extremely lightweight in terms of resources, which makes them widely usable: they can be trained on a standard laptop's CPU in several minutes. This property holds promise to facilitate the development of intent classifiers even without access to large computational resources, which in turn also increases equality and fairness in research \cite{Strubell:2019acl}.

In future work we will port the efficient intent detectors based on dual encoders to other languages, leveraging multilingual pretrained representations \cite{Chidambaram:2019repl}. This work has also empirically validated that there is still ample room for improvement in the intent detection task especially in low-data regimes. Thus, similar to recent work \cite{Upadhyay:2018icassp,Khalil:2019emnlp,Liu:2019emnlp}, we will also investigate how to transfer intent detectors to low-resource target languages in few-shot and zero-shot scenarios. We will also extend the models to handle out-of-scope prediction \cite{larson-etal-2019-evaluation}.

We have released the code and the data sets online at: \\
{\small \url{github.com/PolyAI-LDN/polyai-models}}.

%\clearpage
\bibliography{acl2020}

\begin{thebibliography}{36}
\expandafter\ifx\csname natexlab\endcsname\relax\def\natexlab#1{#1}\fi

\bibitem[{Al{-}Rfou et~al.(2016)Al{-}Rfou, Pickett, Snaider, Sung, Strope, and
  Kurzweil}]{AlRfou:2016arxiv}
Rami Al{-}Rfou, Marc Pickett, Javier Snaider, Yun{-}Hsuan Sung, Brian Strope,
  and Ray Kurzweil. 2016.
\newblock \href {http://arxiv.org/abs/1606.00372} {Conversational contextual
  cues: {T}he case of personalization and history for response ranking}.
\newblock \emph{CoRR}, abs/1606.00372.

\bibitem[{Arase and Tsujii(2019)}]{Arase:2019emnlp}
Yuki Arase and Jun{'}ichi Tsujii. 2019.
\newblock \href {https://www.aclweb.org/anthology/D19-1542} {Transfer
  fine-tuning: {A BERT} case study}.
\newblock In \emph{Proceedings of EMNLP-IJCNLP}, pages 5392--5403.

\bibitem[{Braun et~al.(2017)Braun, Mendez, Matthes, and Langen}]{Braun:17}
Daniel Braun, Adrian~Hernandez Mendez, Florian Matthes, and Manfred Langen.
  2017.
\newblock \href {https://www.aclweb.org/anthology/W17-5522} {Evaluating natural
  language understanding services for conversational question answering
  systems}.
\newblock In \emph{Proceedings of SIGDIAL}, pages 174--185.

\bibitem[{Budzianowski et~al.(2018)Budzianowski, Wen, Tseng, Casanueva, Ultes,
  Ramadan, and Ga\v{s}i\'{c}}]{Budzianowski:2018emnlp}
Pawe{\l} Budzianowski, Tsung-Hsien Wen, Bo-Hsiang Tseng, I{\~{n}}igo Casanueva,
  Stefan Ultes, Osman Ramadan, and Milica Ga\v{s}i\'{c}. 2018.
\newblock \href {http://aclweb.org/anthology/D18-1547} {{MultiWOZ - A}
  large-scale multi-domain wizard-of-oz dataset for task-oriented dialogue
  modelling}.
\newblock In \emph{Proceedings of EMNLP}, pages 5016--5026.

\bibitem[{Cer et~al.(2018)Cer, Yang, Kong, Hua, Limtiaco, John, Constant,
  Guajardo{-}Cespedes, Yuan, Tar, Sung, Strope, and Kurzweil}]{Cer:2018arxiv}
Daniel Cer, Yinfei Yang, Sheng{-}yi Kong, Nan Hua, Nicole Limtiaco, Rhomni~St.
  John, Noah Constant, Mario Guajardo{-}Cespedes, Steve Yuan, Chris Tar,
  Yun{-}Hsuan Sung, Brian Strope, and Ray Kurzweil. 2018.
\newblock \href {http://arxiv.org/abs/1803.11175} {Universal sentence encoder}.
\newblock \emph{CoRR}, abs/1803.11175.

\bibitem[{Chidambaram et~al.(2019)Chidambaram, Yang, Cer, Yuan, Sung, Strope,
  and Kurzweil}]{Chidambaram:2019repl}
Muthuraman Chidambaram, Yinfei Yang, Daniel Cer, Steve Yuan, Yun{-}Hsuan Sung,
  Brian Strope, and Ray Kurzweil. 2019.
\newblock \href {https://www.aclweb.org/anthology/W19-4330/} {Learning
  cross-lingual sentence representations via a multi-task dual-encoder model}.
\newblock In \emph{Proceedings of the 4th Workshop on Representation Learning
  for NLP}, pages 250--259.

\bibitem[{Coucke et~al.(2018)Coucke, Saade, Ball, Bluche, Caulier, Leroy,
  Doumouro, Gisselbrecht, Caltagirone, Lavril et~al.}]{Coucke:18}
Alice Coucke, Alaa Saade, Adrien Ball, Th{\'e}odore Bluche, Alexandre Caulier,
  David Leroy, Cl{\'e}ment Doumouro, Thibault Gisselbrecht, Francesco
  Caltagirone, Thibaut Lavril, et~al. 2018.
\newblock \href {http://arxiv.org/abs/1805.10190} {{Snips Voice Platform: A}n
  embedded spoken language understanding system for private-by-design voice
  interfaces}.
\newblock \emph{arXiv preprint arXiv:1805.10190}, pages 12--16.

\bibitem[{Devlin et~al.(2019)Devlin, Chang, Lee, and
  Toutanova}]{Devlin:2018arxiv}
Jacob Devlin, Ming{-}Wei Chang, Kenton Lee, and Kristina Toutanova. 2019.
\newblock \href {https://www.aclweb.org/anthology/N19-1423} {{BERT:
  P}re-training of deep bidirectional transformers for language understanding}.
\newblock In \emph{Proceedings of NAACL-HLT}, pages 4171--4186.

\bibitem[{El~Asri et~al.(2017)El~Asri, Schulz, Sharma, Zumer, Harris, Fine,
  Mehrotra, and Suleman}]{ElAsri:2017sigdial}
Layla El~Asri, Hannes Schulz, Shikhar Sharma, Jeremie Zumer, Justin Harris,
  Emery Fine, Rahul Mehrotra, and Kaheer Suleman. 2017.
\newblock \href {http://aclweb.org/anthology/W17-5526} {{Frames: A} corpus for
  adding memory to goal-oriented dialogue systems}.
\newblock In \emph{Proceedings of SIGDIAL}, pages 207--219.

\bibitem[{Hemphill et~al.(1990)Hemphill, Godfrey, and
  Doddington}]{Hemphill:1990}
Charles~T. Hemphill, John~J. Godfrey, and George~R. Doddington. 1990.
\newblock {The ATIS Spoken Language Systems Pilot Corpus}.
\newblock In \emph{Proceedings of the Workshop on Speech and Natural Language},
  pages 96--101.

\bibitem[{Henderson et~al.(2019{\natexlab{a}})Henderson, Budzianowski,
  Casanueva, Coope, Gerz, Kumar, Mrk\v{s}i\'{c}, Spithourakis, Su, Vuli\'{c},
  and Wen}]{Henderson:2019arxiv}
Matthew Henderson, Pawel Budzianowski, I{\~{n}}igo Casanueva, Sam Coope,
  Daniela Gerz, Girish Kumar, Nikola Mrk\v{s}i\'{c}, Georgios Spithourakis,
  Pei{-}Hao Su, Ivan Vuli\'{c}, and Tsung{-}Hsien Wen. 2019{\natexlab{a}}.
\newblock \href {http://arxiv.org/abs/1904.06472} {A repository of
  conversational datasets}.
\newblock In \emph{Proceedings of the 1st Workshop on Natural Language
  Processing for Conversational AI}, pages 1--10.

\bibitem[{Henderson et~al.(2019{\natexlab{b}})Henderson, Casanueva,
  Mrk{\v{s}}i{\'c}, Su, Vuli{\'c} et~al.}]{henderson2019convert}
Matthew Henderson, I{\~n}igo Casanueva, Nikola Mrk{\v{s}}i{\'c}, Pei-Hao Su,
  Ivan Vuli{\'c}, et~al. 2019{\natexlab{b}}.
\newblock \href {https://arxiv.org/abs/1911.03688} {Conve{RT: E}fficient and
  accurate conversational representations from transformers}.
\newblock \emph{arXiv preprint arXiv:1911.03688}.

\bibitem[{Henderson et~al.(2019{\natexlab{c}})Henderson, Vuli{\'c}, Gerz,
  Casanueva, Budzianowski, Coope, Spithourakis, Wen, Mrk{\v{s}}i{\'c}, and
  Su}]{Henderson:2019acl}
Matthew Henderson, Ivan Vuli{\'c}, Daniela Gerz, I{\~n}igo Casanueva, Pawe{\l}
  Budzianowski, Sam Coope, Georgios Spithourakis, Tsung-Hsien Wen, Nikola
  Mrk{\v{s}}i{\'c}, and Pei-Hao Su. 2019{\natexlab{c}}.
\newblock \href {https://www.aclweb.org/anthology/P19-1536} {Training neural
  response selection for task-oriented dialogue systems}.
\newblock In \emph{Proceedings of ACL}, pages 5392--5404.

\bibitem[{Humeau et~al.(2020)Humeau, Shuster, Lachaux, and
  Weston}]{Humeau:2019arxiv}
Samuel Humeau, Kurt Shuster, Marie{-}Anne Lachaux, and Jason Weston. 2020.
\newblock \href {http://arxiv.org/abs/1905.01969} {Poly-encoders: {T}ransformer
  architectures and pre-training strategies for fast and accurate
  multi-sentence scoring}.
\newblock In \emph{Proceedings of ICLR}, volume abs/1905.01969.

\bibitem[{Khalil et~al.(2019)Khalil, Kie{\l}czewski, Chouliaras, Keldibek, and
  Versteegh}]{Khalil:2019emnlp}
Talaat Khalil, Kornel Kie{\l}czewski, Georgios~Christos Chouliaras, Amina
  Keldibek, and Maarten Versteegh. 2019.
\newblock \href {https://www.aclweb.org/anthology/D19-1676} {Cross-lingual
  intent classification in a low resource industrial setting}.
\newblock In \emph{Proceedings of EMNLP-IJCNLP}, pages 6418--6423.

\bibitem[{Larson et~al.(2019)Larson, Mahendran, Peper, Clarke, Lee, Hill,
  Kummerfeld, Leach, Laurenzano, Tang, and Mars}]{larson-etal-2019-evaluation}
Stefan Larson, Anish Mahendran, Joseph~J. Peper, Christopher Clarke, Andrew
  Lee, Parker Hill, Jonathan~K. Kummerfeld, Kevin Leach, Michael~A. Laurenzano,
  Lingjia Tang, and Jason Mars. 2019.
\newblock \href {https://www.aclweb.org/anthology/D19-1131} {An evaluation
  dataset for intent classification and out-of-scope prediction}.
\newblock In \emph{Proceedings of EMNLP-IJCNLP}, pages 1311--1316.

\bibitem[{Liu et~al.(2019{\natexlab{a}})Liu, Eshghi, Swietojanski, and
  Rieser}]{Liu:2019iwsds}
Xingkun Liu, Arash Eshghi, Pawel Swietojanski, and Verena Rieser.
  2019{\natexlab{a}}.
\newblock \href {https://arxiv.org/pdf/1903.05566.pdf} {Benchmarking natural
  language understanding services for building conversational agents}.
\newblock In \emph{Proceedings of IWSDS}.

\bibitem[{Liu et~al.(2019{\natexlab{b}})Liu, Ott, Goyal, Du, Joshi, Chen, Levy,
  Lewis, Zettlemoyer, and Stoyanov}]{Liu:2019roberta}
Yinhan Liu, Myle Ott, Naman Goyal, Jingfei Du, Mandar Joshi, Danqi Chen, Omer
  Levy, Mike Lewis, Luke Zettlemoyer, and Veselin Stoyanov. 2019{\natexlab{b}}.
\newblock \href {https://arxiv.org/abs/1907.11692} {{RoBERTa: A} robustly
  optimized {BERT} pretraining approach}.
\newblock \emph{CoRR}, abs/1907.11692.

\bibitem[{Liu et~al.(2019{\natexlab{c}})Liu, Shin, Xu, Winata, Xu, Madotto, and
  Fung}]{Liu:2019emnlp}
Zihan Liu, Jamin Shin, Yan Xu, Genta~Indra Winata, Peng Xu, Andrea Madotto, and
  Pascale Fung. 2019{\natexlab{c}}.
\newblock \href {https://www.aclweb.org/anthology/D19-1129} {Zero-shot
  cross-lingual dialogue systems with transferable latent variables}.
\newblock In \emph{Proceedings of EMNLP-IJCNLP}, pages 1297--1303.

\bibitem[{Maas et~al.(2013)Maas, Hannun, and Ng}]{Maas:2014icml}
Andrew~L. Maas, Awni~Y. Hannun, and Andrew~Y. Ng. 2013.
\newblock \href
  {https://web.stanford.edu/~awni/papers/relu_hybrid_icml2013_final.pdf}
  {Rectifier nonlinearities improve neural network acoustic models}.
\newblock In \emph{Proceedings of ICML}.

\bibitem[{Mehri et~al.(2019)Mehri, Razumovskaia, Zhao, and
  Eskenazi}]{Mehri:2019acl}
Shikib Mehri, Evgeniia Razumovskaia, Tiancheng Zhao, and Maxine Eskenazi. 2019.
\newblock \href {https://www.aclweb.org/anthology/P19-1373} {Pretraining
  methods for dialog context representation learning}.
\newblock In \emph{Proceedings of ACL}, pages 3836--3845.

\bibitem[{Olson et~al.(2018)Olson, Wyner, and Berk}]{Olson:2018nips}
Matthew Olson, Abraham~J. Wyner, and Richard Berk. 2018.
\newblock \href
  {http://papers.nips.cc/paper/7620-modern-neural-networks-generalize-on-small-data-sets}
  {Modern neural networks generalize on small data sets}.
\newblock In \emph{Proceedings of NeurIPS}, pages 3623--3632.

\bibitem[{Peters et~al.(2018)Peters, Neumann, Iyyer, Gardner, Clark, Lee, and
  Zettlemoyer}]{Peters:2018naacl}
Matthew Peters, Mark Neumann, Mohit Iyyer, Matt Gardner, Christopher Clark,
  Kenton Lee, and Luke Zettlemoyer. 2018.
\newblock \href {http://aclweb.org/anthology/N18-1202} {Deep contextualized
  word representations}.
\newblock In \emph{Proceedings of NAACL-HLT}, pages 2227--2237.

\bibitem[{Peters et~al.(2019)Peters, Ruder, and Smith}]{Peters:2019repl}
Matthew~E. Peters, Sebastian Ruder, and Noah~A. Smith. 2019.
\newblock \href {https://www.aclweb.org/anthology/W19-4302} {To tune or not to
  tune? {A}dapting pretrained representations to diverse tasks}.
\newblock In \emph{Proceedings of the 4th Workshop on Representation Learning
  for NLP}, pages 7--14.

\bibitem[{Radford et~al.(2019)Radford, Wu, Child, Luan, Amodei, and
  Sutskever}]{radford2019language}
Alec Radford, Jeffrey Wu, Rewon Child, David Luan, Dario Amodei, and Ilya
  Sutskever. 2019.
\newblock \href
  {https://d4mucfpksywv.cloudfront.net/better-language-models/language_models_are_unsupervised_multitask_learners.pdf}
  {Language models are unsupervised multitask learners}.
\newblock \emph{OpenAI Blog}, 1(8).

\bibitem[{Rastogi et~al.(2019)Rastogi, Zang, Sunkara, Gupta, and
  Khaitan}]{rastogi2019towards}
Abhinav Rastogi, Xiaoxue Zang, Srinivas Sunkara, Raghav Gupta, and Pranav
  Khaitan. 2019.
\newblock \href {https://arxiv.org/pdf/1909.05855.pdf} {Towards scalable
  multi-domain conversational agents: The schema-guided dialogue dataset}.
\newblock \emph{arXiv preprint arXiv:1909.05855}.

\bibitem[{Ruder et~al.(2019)Ruder, Peters, Swayamdipta, and
  Wolf}]{Ruder:2019transfer}
Sebastian Ruder, Matthew~E. Peters, Swabha Swayamdipta, and Thomas Wolf. 2019.
\newblock \href {https://www.aclweb.org/anthology/N19-5004} {Transfer learning
  in natural language processing}.
\newblock In \emph{Proceedings of NAACL-HLT: Tutorials}, pages 15--18.

\bibitem[{Sanh et~al.(2019)Sanh, Debut, Chaumond, and Wolf}]{Sanh:2019arxiv}
Victor Sanh, Lysandre Debut, Julien Chaumond, and Thomas Wolf. 2019.
\newblock \href {https://arxiv.org/abs/1910.01108} {{DistilBERT}, a distilled
  version of {BERT: S}maller, faster, cheaper and lighter}.
\newblock \emph{CoRR}, abs/1910.01108.

\bibitem[{Srivastava et~al.(2014)Srivastava, Hinton, Krizhevsky, Sutskever, and
  Salakhutdinov}]{Srivastava:2014jmlr}
Nitish Srivastava, Geoffrey~E. Hinton, Alex Krizhevsky, Ilya Sutskever, and
  Ruslan Salakhutdinov. 2014.
\newblock \href {http://dl.acm.org/citation.cfm?id=2670313} {Dropout: {A}
  simple way to prevent neural networks from overfitting}.
\newblock \emph{Journal of Machine Learning Research}, 15(1):1929--1958.

\bibitem[{Strubell et~al.(2019)Strubell, Ganesh, and
  McCallum}]{Strubell:2019acl}
Emma Strubell, Ananya Ganesh, and Andrew McCallum. 2019.
\newblock \href {https://www.aclweb.org/anthology/P19-1355} {Energy and policy
  considerations for deep learning in {NLP}}.
\newblock In \emph{Proceedings of ACL}, pages 3645--3650.

\bibitem[{Upadhyay et~al.(2018)Upadhyay, Faruqui, T{\"{u}}r,
  Hakkani{-}T{\"{u}}r, and Heck}]{Upadhyay:2018icassp}
Shyam Upadhyay, Manaal Faruqui, G{\"{o}}khan T{\"{u}}r, Dilek
  Hakkani{-}T{\"{u}}r, and Larry~P. Heck. 2018.
\newblock \href {https://doi.org/10.1109/ICASSP.2018.8461905} {(almost)
  zero-shot cross-lingual spoken language understanding}.
\newblock In \emph{Proceedings of ICASSP}, pages 6034--6038.

\bibitem[{Wen et~al.(2017)Wen, Vandyke, Mrk{\v{s}}i\'c, Ga{\v{s}}i\'c,
  M.~Rojas-Barahona, Su, Ultes, and Young}]{Wen:17}
Tsung-Hsien Wen, David Vandyke, Nikola Mrk{\v{s}}i\'c, Milica Ga{\v{s}}i\'c,
  Lina M.~Rojas-Barahona, Pei-Hao Su, Stefan Ultes, and Steve Young. 2017.
\newblock \href {http://www.aclweb.org/anthology/E17-1042} {A network-based
  end-to-end trainable task-oriented dialogue system}.
\newblock In \emph{Proceedings of EACL}, pages 438--449.

\bibitem[{Williams(2012)}]{Williams:2012b}
Jason Williams. 2012.
\newblock {A critical analysis of two statistical spoken dialog systems in
  public use}.
\newblock In \emph{Proceedings of SLT}.

\bibitem[{Xu et~al.(2017)Xu, Liu, Guo, Sinha, and Akkiraju}]{Xu:2017chi}
Anbang Xu, Zhe Liu, Yufan Guo, Vibha Sinha, and Rama Akkiraju. 2017.
\newblock \href {http://doi.acm.org/10.1145/3025453.3025496} {A new chatbot for
  customer service on social media}.
\newblock In \emph{Proceedings of the 2017 CHI Conference on Human Factors in
  Computing Systems}, pages 3506--3510.

\bibitem[{Yang et~al.(2019)Yang, Cer, Ahmad, Guo, Law, Constant, {\'{A}}brego,
  Yuan, Tar, Sung, Strope, and Kurzweil}]{Yang:2019multiuse}
Yinfei Yang, Daniel Cer, Amin Ahmad, Mandy Guo, Jax Law, Noah Constant,
  Gustavo~Hern{\'{a}}ndez {\'{A}}brego, Steve Yuan, Chris Tar, Yun{-}Hsuan
  Sung, Brian Strope, and Ray Kurzweil. 2019.
\newblock \href {http://arxiv.org/abs/1907.04307} {Multilingual universal
  sentence encoder for semantic retrieval}.
\newblock \emph{CoRR}, abs/1907.04307.

\bibitem[{Zafrir et~al.(2019)Zafrir, Boudoukh, Izsak, and
  Wasserblat}]{Zafrir:2019arxiv}
Ofir Zafrir, Guy Boudoukh, Peter Izsak, and Moshe Wasserblat. 2019.
\newblock \href {https://arxiv.org/abs/1910.06188} {{Q8BERT: Q}uantized 8bit
  {BERT}}.
\newblock \emph{CoRR}, abs/1910.06188.

\end{thebibliography}
\bibliographystyle{acl_natbib}

%\appendix

%%\section{Appendix 1}
%%\label{sec:appendix}
%%\dan{Do we need an appendix?}

\end{document}